\documentclass[format=sigconf, natbib=false, nonacm=true, screen=true]{acmart}

\usepackage[T1]{fontenc}
\usepackage[utf8]{inputenc}
\usepackage{csquotes}
\usepackage[inline]{enumitem}

\RequirePackage[abbreviate=true, dateabbrev=true, isbn=false, doi=true, url=false, eprint=false, backend=biber, bibencoding=utf8, style=ACM-Reference-Format]{biblatex}

\copyrightyear{2020}
\acmYear{2020}
\setcopyright{rightsretained}
\acmConference[HRI '20 Companion]{Companion of the 2020 ACM/IEEEInternational Conference on Human-Robot Interaction}{March 23--26,2020}{Cambridge, United Kingdom}
\acmBooktitle{Companion of the 2020 ACM/IEEE International Conference
  on Human-Robot Interaction (HRI '20 Companion), March 23--26, 2020,Cambridge, United Kingdom}
\acmPrice{}
\acmDOI{10.1145/3371382.3378257}
\acmISBN{978-1-4503-7057-8/20/03}

\acmSubmissionID{lbr1046}

\newcommand{\term}[1]{{\textit{#1}}}

\begin{document}
\fancyhead{}

%%
%% The "title" command has an optional parameter,
%% allowing the author to define a "short title" to be used in page headers.
\title{Using Social Robots to Teach Language Skills to Immigrant Children in an Oslo City District}

%%
%% The "author" command and its associated commands are used to define
%% the authors and their affiliations.
%% Of note is the shared affiliation of the first two authors, and the
%% "authornote" and "authornotemark" commands
%% used to denote shared contribution to the research.
\author{Trenton Schulz}
\email{trenton.schulz@nr.no}
\orcid{0000-0001-6217-758X}
\author{Till Halbach}
\email{till.halback@nr.no}
\author{Ivar Solheim}
\email{ivar.solheim@nr.no}
\affiliation{%
  \institution{Norwegian Computing Center}
  \streetaddress{Gaustadalleen 23a}
  \city{Oslo}
  \country{Norway}
  \postcode{NO-0373}
}

%%
%% By default, the full list of authors will be used in the page
%% headers. Often, this list is too long, and will overlap
%% other information printed in the page headers. This command allows
%% the author to define a more concise list
%% of authors' names for this purpose.
% \renewcommand{\shortauthors}{Schulz et al.}

%%
%% The abstract is a short summary of the work to be presented in the
%% article.
\begin{abstract}
  Social robots have been shown to help in language education for
  children. This can be good aid for immigrant children that need
  additional help to learn a second language their parents do not
  understand to attend school. We present the setup for a long-term
  study that is being carried out in blinded to aid immigrant children
  with poor skills in the Norwegian language to improve their
  vocabulary. This includes additional tools to help parents follow
  along and provide additional help at home.
\end{abstract}

%%
%% The code below is generated by the tool at http://dl.acm.org/ccs.cfm.
%% Please copy and paste the code instead of the example below.
%%
\begin{CCSXML}
<ccs2012>
<concept>
<concept_id>10003120.10003121.10003124.10010870</concept_id>
<concept_desc>Human-centered computing~Natural language interfaces</concept_desc>
<concept_significance>500</concept_significance>
</concept>
<concept>
<concept_id>10003120.10003121.10003122.10003334</concept_id>
<concept_desc>Human-centered computing~User studies</concept_desc>
<concept_significance>300</concept_significance>
</concept>
<concept>
<concept_id>10010520.10010553.10010554</concept_id>
<concept_desc>Computer systems organization~Robotics</concept_desc>
<concept_significance>500</concept_significance>
</concept>
</ccs2012>
\end{CCSXML}

\ccsdesc[500]{Human-centered computing~Natural language interfaces}
\ccsdesc[300]{Human-centered computing~User studies}
\ccsdesc[500]{Computer systems organization~Robotics}

%%
%% Keywords. The author(s) should pick words that accurately describe
%% the work being presented. Separate the keywords with commas.
\keywords{social robots; language education; children; teaching}

\maketitle

\section{Introduction}

Children with an immigrant background, including refugees, may have
difficulties learning and speaking the local language. This can be a
real issue when the children begin in school and are expected to have
a certain language proficiency. The situation can put a strain on the school
system and teachers to provide the necessary basic language skills
while also teaching the rest of the children. The problem is faced by
several city districts in Oslo where up to 40\% of the children who
enter primary school need additional teaching in the Norwegian language.
Children face further difficulties to improve their Norwegian if
it is not spoken at home. Children may lose their
motivation to stay with the studies and fall further behind their
peers. Using social robots to help in the language training may
offload some of the teaching load and keep the children motivated.

\begin{figure}[tb]
  \includegraphics[width=1.\linewidth]{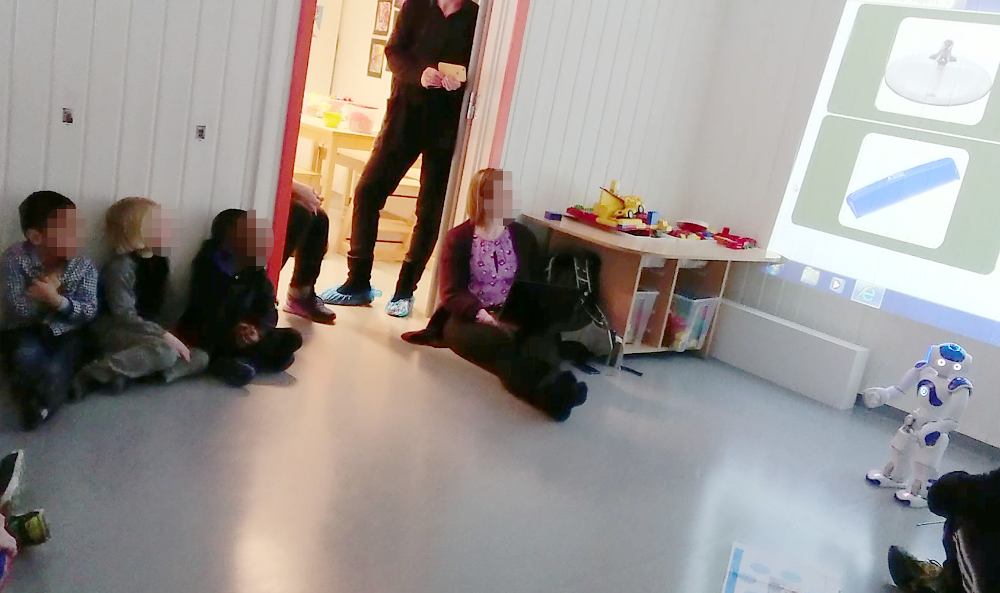}
  \caption{Nao interacting with a small group of children in a care center
    context with staff present. Pictures of the terms to be learned are
    projected on the wall behind the robot.}
  \label{fig:bhg}
\end{figure}

We are designing a program to use a social robot to improve
immigrant children's ability to learn Norwegian as part of
the language education in daycare centers (kindergartens). The
program is designed with the help of parents and employees from
a city district in Oslo. This paper documents the work
that has been done so far, and how the study will be carried out over
the next year.

\section{Related work}

Social robots have been used for teaching language in a variety of
situations. The L2TOR project introduced a possible design for using a
robot and tablet to teach children a second language
\parencite{belpaemeL2TORSecondLanguage2015}. Experiences from this
study and other have lead to guidelines for teaching second language
to children \parencite{belpaemeGuidelinesDesigningSocial2018}. These
guidelines highlight factors to consider such as the modality and if
the robot acts as a peer, a teacher, or provides no feedback.

A recent large scale study compared how well 194 Dutch primary school
children could learn English with a Nao versus a tablet application
\parencite{vogtSecondLanguageTutoring2019}. The study showed that
children could retain the language they had learned from the robot at
the same level they could with a tablet and that iconic gestures from the
robot did not seem to effect on how well the children learned. Our
work differs in that we are teaching slightly younger immigrant
children Norwegian so they can better participate in schools and their
education.

Beyond language education, robots were programmed to show empathy to
see how it affected children playing a game to learn about
sustainability issues
\parencite{alves-oliveiraEmpathicRobotGroup2019}. Others have also
used robots and language to explore children's trust of a robot
\parencite{geiskkovitchWhatThatNot2019}.

This study draws inspiration from our earlier pilot study using social
robots and tablets to teach Norwegian to children in daycare centers
\parencite{fuglerudUseSocialRobots2018}. The pilot consisted of the
robot together with a teacher, a group of children, and a tablet. The
pilot showed that the children needed personalized one-on-one
attention that could be provided by the robot, and that the speech
recognition for Norwegian (especially for children, a general speech
recognition problem \parencite{kennedyChildSpeechRecognition2017})
must be improved. In addition, we are using guidelines from
\parencite{belpaemeGuidelinesDesigningSocial2018}.

\section{Small-scale study and interviews}

To help children before they enter primary school, we targeted day
care centers in the Grorud district of Oslo. Most children in the
district attend daycare centers from ages one to five. The pedagogic
staff of the centers have developed a language program,
\term{språkdusj} (language shower), to aid children in expanding their
Norwegian vocabulary. Currently, the program is implemented as printed
pictures of things for which the proper term is to be learned. The
basic principle has children presented with a corresponding visual and
the question \enquote{What do you see here?} The children have to
recall the correct term from their memory and its pronunciation by
saying the word aloud. To reduce issues with the robot recognizing
children's speech, the pedagogic staff would assist the children in
the process and determine if the answer was correct.

A small-scale study was conducted in Autumn 2018 in selected daycare
centers. The trial involved a prototype using the Nao robot
\parencite{softbankrobotics} that stood in front of approximately 15
children (Figure~\ref{fig:bhg}). The Nao was linked to a web
application that was projected on the wall and showed pictures of the
terms to be learned (e.g., trousers, chair, fork, showers) one at a
time. The robot asked what the children saw, processed the children's
answer in its speech recognition engine, and determined the answer's
correctness.

The study showed promising results in
children's motivation as compared to the language program without a
robot. From the study, the robot's main advantages appeared to be its
attractive, human-like appearance; its not-threatening size for
children; speech synthesis and recognition; and the ability to move,
gesture, and dance, including support for light, as well as sound and
audio effects.

We conducted several interviews and workshops with pedagogues and
other daycare center staff to find out how a robot could be part of
their language program. We also interviewed some of the children's
parents to understand the parents' Norwegian abilities and if children
spoke (and learned) Norwegian or other languages at home. Overall, twelve people
were interviewed. Suggestions included having the children in dialog
with a robot and including game elements that would track each child's
progress to help motivate the children. The interviews also revealed
it was necessary to include parents in the language teaching since
they can help maintain the children's motivation at home, and they may
also benefit from learning Norwegian themselves.

\section{Large-scale Study Design}

We used the suggestions from the interviews and workshops to build on
the previous study by developing a digitalized version of the language
program, and expanded it to include adults. The program includes a
Nao, a tablet app, and a mobile app. The tablet app is a prompt and
starting point for lessons with Nao. For instance, in one interaction,
the app shows a picture with three socks. Nao asks the children what
is the pattern of the sock in the middle? Without Nao, the app shows
four answer alternatives in a multiple-choice manner. To maintain
motivation, Nao will provide varied supporting feedback and give
rewards in form of oral acknowledgments, dances, and suitable light
and sound effects. The program also adds a mobile application that
targets the children's parents at home The mobile app helps to keep
track of a single child's or group of children's progress and can be
used to have discussions about language at home. Currently, the app
contains the entire language program and all instructions for the
robot. The robot instructions is an add-on available only in the
daycare centers. If the Nao is available, it connects to the app and
coordinates the walk-through of the language program in a joint
manner.

The study will begin in Spring 2020. The main objectives are
\begin{enumerate*}[label=(\textit{\arabic*})]
\item verify the technology, pedagogical, and gamification concepts;
\item measure gains in children's permanent vocabulary using Nao and the accompanying apps
  compared to the traditional language program.
\end{enumerate*}
The goal is the children will be better prepared to meet the language
vocabulary requirements of the primary school, with a further positive
impact on work life and social life as well.

The study will be longitudinal and use a between-subjects design.  Some
district daycare centers will have access to the digitalized
solution, others not. The study will start at one daycare center to control for
problems and added to other centers as the system stabilizes.
Ideally, the centers' employees will control the pace.

We will compare the language
development of the children that use the program versus those that do
not. We plan to test the children's language skills at the
beginning of the study, during the study, when it terminates, and 6
months after the trials' termination. This involves testing the words
that have been taught in the robot sessions to measure language
skills.

\section{Discussion and Conclusion}

Involving parents and daycare employees has helped create a novel
system for language training of immigrant children and maintaining
their motivation. There are potential issues, particularly in speech
synthesis and speech recognition for children, that may cause the
system to fail. We are continually investigating these issues and hope
to have them addressed before or during its use in the first daycare
center. It will also be interesting to see how the robot becomes
part of each daycare center's education program. Our study design
should provide good evidence to whether a social robot can be an
important tool for immigrant children to learn the Norwegian language before
entering primary school.

%%
%% The acknowledgments section is defined using the "acks" environment
%% (and NOT an unnumbered section). This ensures the proper
%% identification of the section in the article metadata, and the
%% consistent spelling of the heading.
\begin{acks}
Thanks to the parents, children, and day car workers at Grorud City District in Oslo, master students in Oslo Metropolitan University, and Innocom AS for help in realizing this project.
\end{acks}

%%
%% The next two lines define the bibliography style to be used, and
%% the bibliography file.
\printbibliography

\end{document}